\documentclass[10pt, a4paper]{article}
\usepackage{lrec2000}
\usepackage{alltt, epsfig, url}
\def\smtt#1{{\small\texttt{#1}}}

\def\smurl#1{[{\small\url{#1}}]}

\def\arXivhack{\vspace{-6pt}}

\title{TableTrans, MultiTrans, InterTrans and TreeTrans:\\
  Diverse Tools Built on the Annotation Graph Toolkit}

\name{
Steven Bird,
Kazuaki Maeda,
Xiaoyi Ma,
Haejoong Lee,
Beth Randall, and
Salim Zayat
}

\address{Linguistic Data Consortium, University of Pennsylvania\\
3615 Market Street, Suite 200, Philadelphia, PA 19104-2608, USA\\
\{sb, maeda, xma, haejoong, brandall, zayats\}@ldc.upenn.edu
}

\abstract{
Four diverse tools built on the Annotation Graph Toolkit are
described.  Each tool associates linguistic codes and structures with
time-series data.  All are based on the same software
library and tool architecture.  TableTrans is for observational
coding, using a spreadsheet whose rows are aligned to a signal.
MultiTrans is for transcribing multi-party communicative interactions
recorded using multi-channel signals.
InterTrans is for creating interlinear text aligned to audio.
TreeTrans is for creating and manipulating syntactic trees.
This work demonstrates that the development of diverse tools and
re-use of software components is greatly facilitated by a common
high-level application programming interface for representing the data
and managing input/output, together with a common architecture for
managing the interaction of multiple components.
}

\begin{document}

\maketitleabstract

\section{Introduction}

Annotation graphs provide an efficient and expressive data model for
linguistic annotations of time-series data.  We have developed a
complete open-source software infrastructure supporting the rapid
development of tools for transcribing and annotating time-series data,
the Annotation Graph Toolkit (AGTK).  This general-purpose
infrastructure uses annotation graphs as the underlying model and
allows developers to quickly create special-purpose annotation tools
using common components.  An application programming interface, an
input/output library supporting a variety of annotation file formats,
and several graphical user interfaces have been developed.

This paper describes four annotation tools based on AGTK which have
been developed at the Linguistic Data Consortium.  TableTrans is for
observational coding, using a spreadsheet whose rows are aligned to a
signal.  MultiTrans is for transcribing multi-party communicative
interactions recorded using multi-channel signals.  InterTrans is for
creating interlinear text aligned to audio.  TreeTrans is for creating
and manipulating syntactic trees.
The toolkit and tools are distributed
under an open source software license.
\arXivhack

\section{The Annotation Graph Toolkit (AGTK)}

An annotation graph \cite{BirdLiberman01} is a directed acyclic graph where
edges are labeled with fielded records, and nodes are (optionally) labeled
with time offsets.  The annotation graph model, a generalization of the
Tipster model used in text retrieval \cite{Grishman96}, is capable of
representing virtually all types of linguistic annotation (e.g. phonetic,
orthographic, part-of-speech, syntactic, discourse, intonational).  This
development has opened up an interesting range of new possibilities for
creation, maintenance and search, and has lead to new annotation tools with
applicability across the text, audio and video modalities.

The Annotation Graph Toolkit (AGTK) is a collection of software for
the development of annotation tools, instantiating the annotation graph
model.  AGTK includes application programming interfaces (APIs) for
manipulating annotation graph data and importing data from other formats,
wrappers for scripting languages, graphical user interface (GUI)
components specialized for annotation tasks, and demonstration applications.

\newcite{lrec-devel} describes the toolkit in detail.  
\newcite{lrec-db} describes collaborative annotation using a shared
database server.
\arXivhack

\section{Use of Third-party Software}

AGTK depends on a variety of third-party software, as detailed below.

\subsection{Scripting languages: Python and Tcl}

AGTK provides interfaces to its libraries for both Tcl and Python.
Tcl is a scripting language with a native GUI toolkit called Tk.
Python is a newer object-oriented scripting language.  The early
AGTK applications were developed using Tcl,
but more recent applications have been developed in Python.
Both sets of applications depend on the Tk GUI library.

\subsection{Audio waveform display: WaveSurfer}

WaveSurfer \cite{Sjolander00} is open source software for displaying and
manipulating audio data, developed by K\r{a}re Sj\"{o}lander and
Jonas Beskow of KTH.  WaveSurfer requires Snack, also developed by
Sj\"{o}lander.  WaveSurfer is written in Tcl/Tk and its widget,
called \emph{wsurf}, can be embedded in any Tcl/Tk application.
There is a Python/Tkinter interface for wsurf that permits it to
be embedded in a Python/Tkinter application.

\begin{figure*}[ht]
  \begin{center}
    \leavevmode
    \epsfig{figure=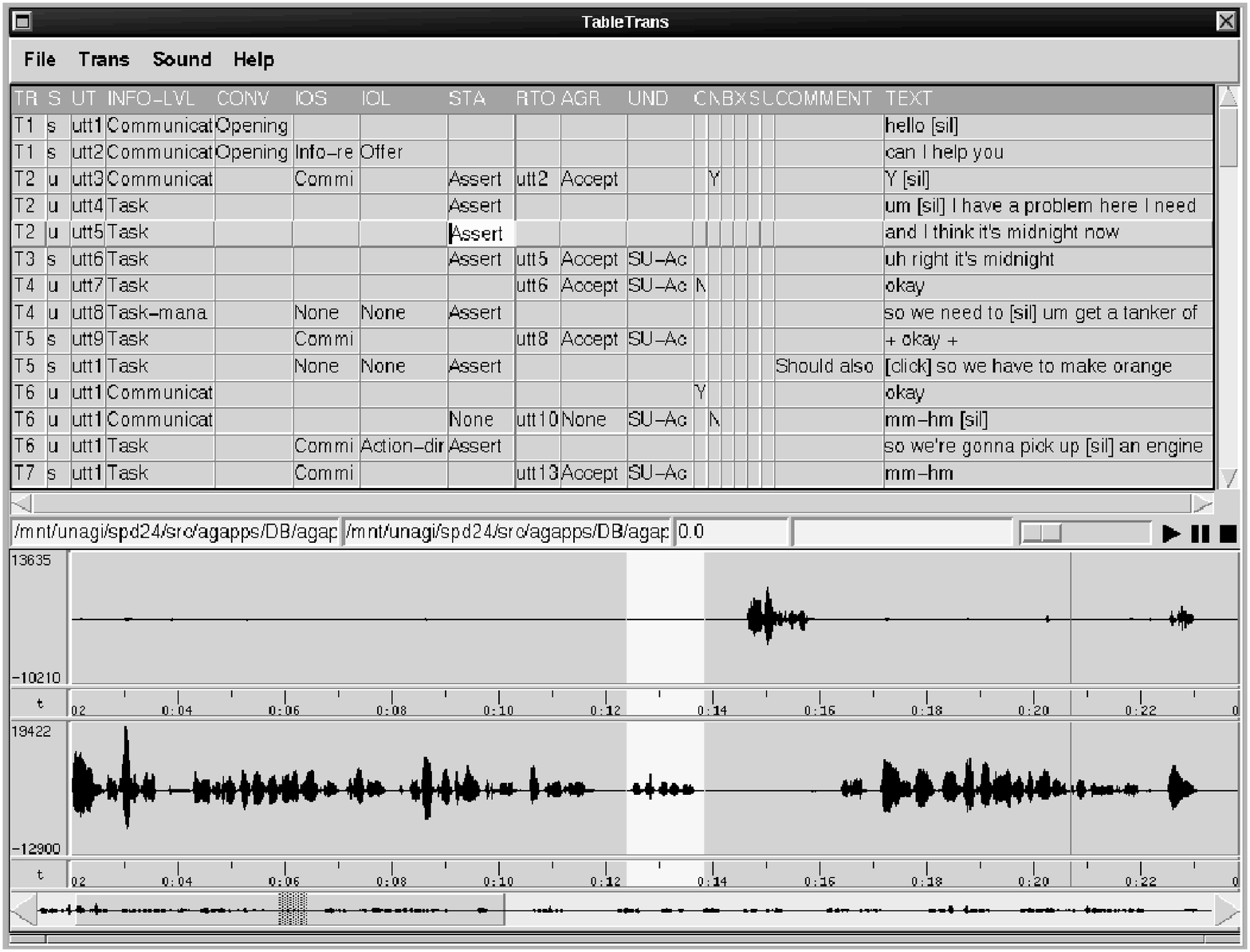, width=0.95\linewidth}
  \end{center}
  \caption{TableTrans With WaveSurfer}
  \label{fig:TableTrans}
\end{figure*}

\subsection{Video display: QuickTime}

QuickTime Tcl \smurl{http://hem.fyristorg.com/matben/qt/} can be embedded
in a Tcl/Tk application.  We have written a simple wrapper in order to embed
QuickTime Tcl in a Python/Tkinter application.  QuickTime Tcl requires the 
QuickTime player distributed by Apple Corporation.
\smurl{http://www.apple.com/quicktime/download/}. 
\arXivhack

\section{TableTrans}

TableTrans is a spreadsheet-based annotation tool for audio signals.
It was primarily intended for linguistic annotations, but it can be
used for many kinds of observational coding, such as widely practiced
in ethology. The tool allows users to annotate regions of signal with
arbitrary features, such as speaker identifiers, utterance identifiers
and transcriptions. Each row of the spreadsheet corresponds to a region of
the signal.  Each column corresponds to a user-defined feature, and
has a user-defined width.  For example, one could define the width of
5 characters for the speaker identifier, and the width of 40 for the
transcription.  One could also choose to define ten or more columns as
shown below, and use it for more structured coding.

\begin{figure}[htbp]
  \begin{center}
    \leavevmode
    \epsfig{figure=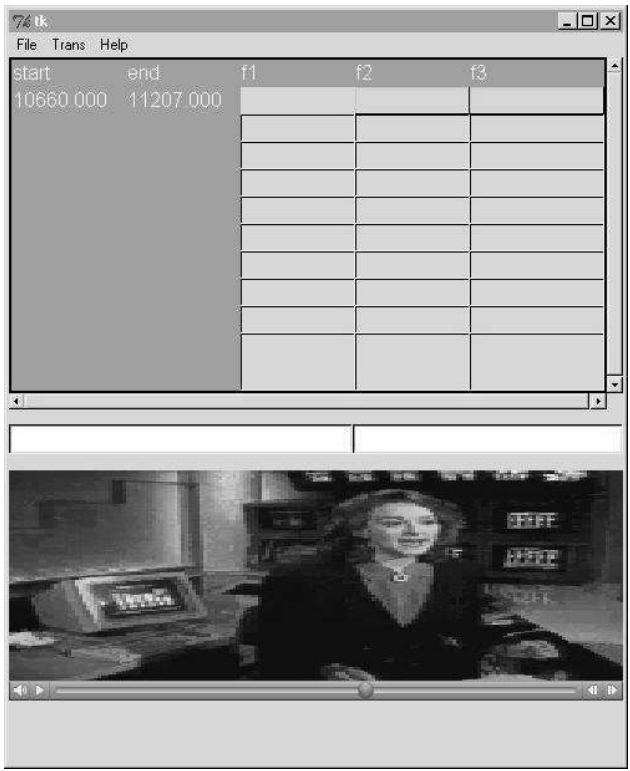, width=1.00\linewidth}
  \end{center}
  \caption{TableTrans With the QuickTime Player}
  \label{fig:TableTransVideo}
\end{figure}

Figure~\ref{fig:TableTrans} shows TableTrans with a waveform display based
on WaveSurfer.
Figure~\ref{fig:TableTransVideo} shows TableTrans with a video module based
on QuickTime and QuickTimeTcl.  


\paragraph{Setting Current Region}

Swiping a region in the waveform display while holding down the
left-mouse button highlights the region.  This region is called
the \emph{current region}.  In the video version of TableTrans, the
current region can be set with two operations: pressing
\emph{control-s} at the beginning of the region, and pressing
\emph{control-e} at the end of the region.

\paragraph{Create Annotation}

Pressing the \emph{Return} key inserts a blank annotation row in
the spreadsheet. If the current region of the waveform is chosen, the start
and end times of the current region are inserted.

\paragraph{Delete Current Annotation}

Pressing \emph{Control-d} deletes the currently
highlighted annotation (the \emph{current annotation}).

\paragraph{Update Current Region}

Pressing \emph{Control-g} updates the start and end
times of the current annotation using the current region in the waveform.

\paragraph{Navigate in Spreadsheet}

The \emph{Tab} key moves the cursor one cell to the right, if possible.
The right, left, up, and down arrows move to the neighboring cells in the
corresponding directions.

\paragraph{Navigate Within Current Cell}

The \emph{Shift-Right} and \emph{Shift-Left} keys move the
insertion point within the current cell.

\paragraph{Toggle Play and Stop}

The \emph{F1} key toggles playback and pausing of the recording. If the
current region is chosen, it plays the current region. If a single
point is selected in the waveform, \emph{F1} starts playback from the point,
and annotations corresponding to the audio cursor are
highlighted (aligned playback).

\paragraph{Sort Annotations}

Double-clicking the left mouse button on a feature name (a column heading)
of the spreadsheet sorts the annotations according to the values in the
column. For example, double clicking on the cell titled 'f1' will sort all
the annotations according to the values in the column. Sorting according to
start and end times can also be done from the menu items,
\smtt{Trans->Sort->Sort Annotations by Start Time} and
\smtt{Trans->Sort->Sort Annotations by End Time}.

\paragraph{Find}

The menu item \smtt{Trans->Find} or the \emph{Control-f} key combination
brings up a dialog window to search a string in the spreadsheet. If a
matching string is found in the cell it is highlighted and the
annotation row becomes the current annotation.

\paragraph{Control View of Annotations}

The menu item \smtt{Trans->View->Show Select Rows} lets the user specify a
feature and a value so that only the annotations having this value as the
feature will be displayed.  The other menu items in \smtt{Trans->View} are
for displaying all annotations and hiding all annotations in the
spreadsheet.

\paragraph{Open Sound/Movie File}

The menu item \smtt{File->Open Sound File} loads a sound file into the
waveform panel. It automatically adjusts the number of waveforms
according to the number of channels in the sound file. All sound file
formats supported by WaveSurfer are supported.  In the video version of
TableTrans, the \smtt{File->Open Movie File} will open a movie file.

\paragraph{Open Annotation File}

The menu item \smtt{File->Open Annotation File} provides the users with
options to load an annotation file in the following formats: XML (AIF),
Table Format (csv, etc.) and LCF (the LDC Callhome Format).

The application opens appropriate dialog windows for each format. First,
the user is prompted for the file.  Then, for a Table format, a window
for specifying names of features and a delimiter is opened.  The names
specified here are used as feature names and column headers. 

\paragraph{Save Annotations in File}

If a file name and a format type have been already chosen, the menu item
\smtt{File->Save} will save all the annotations in the file. If they have
not be chosen, the user needs to use one of the 
\smtt{File->Save Annotations As} menu items.

\paragraph{Save Current Sound Region in File}

The menu item \smtt{Sound->Save Current Region} in \smtt{File}
lets the user save the current region of the sound data into a file.

\paragraph{Column configurations}

Column headers and width may be changed interactively, or may be specified
in a configuration file.    

\paragraph{Database support}

\begin{figure}[htbp]
  \begin{center}
    \leavevmode
    \epsfig{figure=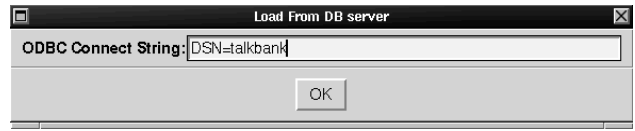, width=1.00\linewidth}
  \end{center}
  \caption{ODBC connect string dialog}
  \label{fig:connstr}
\end{figure}

TableTrans supports access to the database component of AGTK.
Figure~\ref{fig:connstr} shows the dialog window for entering a ODBC
connect string.  Table~\ref{tab:connectstr} shows some of the parameters
used in a connect string.  For a complete list, please see
\smurl{http://www.mysql.com/doc/M/y/MyODBC_connect_parameters.html}.

\begin{table}[htbp]
  \begin{center}
    \leavevmode
{\footnotesize
\begin{tabular}{|l|l|} \hline

DSN & Registered ODBC Data Source Name. \\ \hline
SERVER & The hostname of the database server. \\ \hline
UID & User name as established on the server. \\
& SQL Server this is the logon name. \\ \hline
PWD & Password that corresponds with the logon name.\\ \hline
DATABASE & Database to connect to. If not given, DSN is used.\\ \hline
\end{tabular}
}
\end{center}
  \caption{ODBC Connect String Parameters}
  \label{tab:connectstr}
\end{table}
\arXivhack

\section{MultiTrans}

\begin{figure*}[htbp]
  \begin{center}
    \leavevmode
    \epsfig{figure=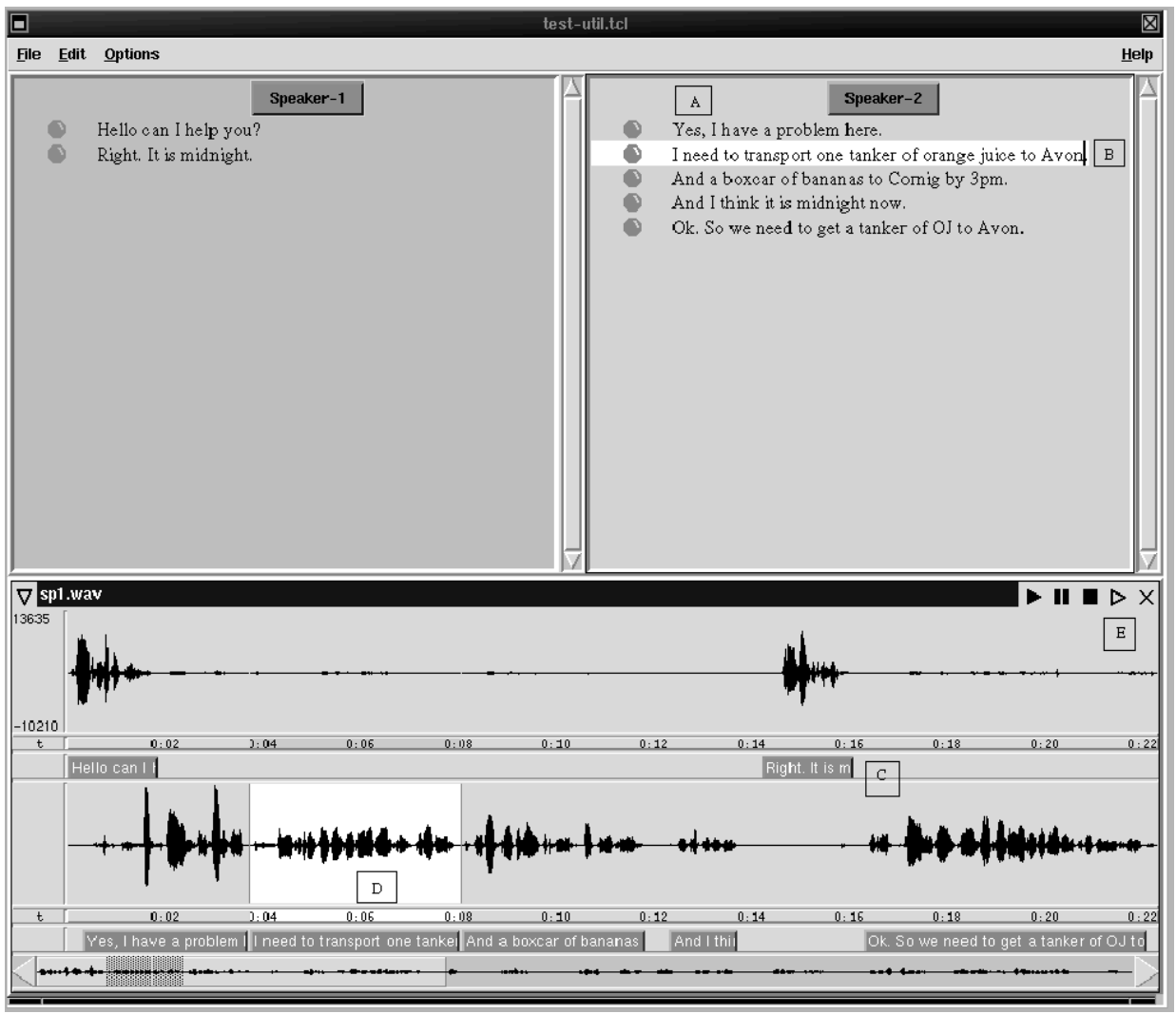, width=0.95\linewidth}
  \end{center}
  \caption{MultiTrans, for Transcribing Multi-Party Conversation}
  \label{fig:MultiTrans}
\end{figure*}

MultiTrans is a transcription tool for transcribing multi-party
conversations in multi-channel audio signals.  The user interface is
similar to Transcriber \cite{Barras01}, but MultiTrans has one
transcription panel corresponding to each channel in the signal.

Figure~\ref{fig:MultiTrans} contains a screenshot of MultiTrans with a
two-channel audio signal.  The left transcription panel corresponds to the
first channel in the audio signal, and the right transcription panel
corresponds to the second channel.  The boxes labeled from A to E in the
figure illustrates the following points.

\begin{itemize}
\item[A:]  The text panel for speaker 2. The channel associated with
speaker 2 is the current channel. 
\item[B:] This is the second annotation for speaker 2. It is also the
current annotation, and is highlighted. 
\item[C:] A segment for speaker 1. This shows some of the transcript
for this annotation and the associated region of the waveform.
\item[D:] The highlighted region of the waveform for the second
annotation. This is the current annotation and its entire waveform region is
highlighted. 
\item[E:] The hollow play button. This button will play the current
channel only, muting other speech channels. 
\end{itemize}

There are two ways to create an annotation, both are explained below. All
annotations are inserted into the text panel, sorted by their starting
position. 

\paragraph{Create Annotation (Specific)} 

The first way to create an annotation is to explicitly highlight a region
in the waveform and press the \emph{Return} key. This will create a bullet
in the appropriate text panel that designates the created annotation.
Also, a region is created below the waveform itself to designate the
created annotation. Only once an annotation is created can the
transcription of that region begin.

\paragraph{Create Annotation (Non-Specific)} 

Another way to create an annotation is during speech playback. When the
playback of speech is begun the user may press the \emph{Return} key to
insert an anchor in the current channel. When \emph{Return} is pressed a
small black bar will appear below the waveform. This designates the
starting position of the current annotation. When \emph{Return} is pressed a second
time the end anchor for the annotation is inserted and the annotation is
created. Ending speech playback destroys any start anchors that do not yet
have an associated end anchor.
 
\paragraph{Delete Current Annotation}
 
An annotation can be deleted using \emph{Control-d}. To
delete an annotation one must first select the annotation to be deleted.
The current annotation can easily be distinguished from others because
its transcription and waveform regions are highlighted. When an
annotation is deleted its transcription and its association with
the waveform region are also deleted.

\paragraph{Change Current Annotation}
 
Once an annotation is created its region in the waveform can be changed
without deleting the entire annotation. This is done by selecting the desired
annotation (with a click in the text panel or segment below the waveform),
and moving either the end point or start point of the annotation and pressing
the \emph{Return} key to register the change.

\paragraph{Split Current Annotation}

Large annotations can be split into smaller annotations using the split
current annotation command. First, the area in the text transcription where
the split is to occur is selected. Next, the area in the waveform where the
split is to occur is selected. Finally, the \emph{Return} key is pressed
and the old annotation is split into two new annotations, each associated
with a different waveform region.

\paragraph{Join Current Annotation}

Joining an annotation is the opposite of splitting an annotation. This is
done by selecting an annotation and pressing the \emph{Shift-BackSpace} key
combination. This will merge the currently selected annotation region and
transcription with the one that occurs immediately before it.

\paragraph{Squeeze Current Annotation}
 
When an annotation is squeezed its starting boundary is pushed to the
ending boundary of the previous annotation. This is used when annotations
are desired to be separate, but one is to begin as soon as the other ends.
This is done by selecting an annotation and pressing the
\emph{Control-Shift-BackSpace} key combination.

\paragraph{Toggle Speech Playback}

There are several ways to begin speech playback. Pressing the \emph{Tab}
key will toggle speech playback of the current annotation, or the entire
speech file if there is not a current annotation. Playback can also be
initiated by pressing the solid play button in the waveform panel.
Either of these commands will play all channels in the speech file.
Pressing the hollow play button located in the waveform panel will play
the current channel only, muting all other channels. This is useful when
there are several channels that make speaker distinction difficult.
\arXivhack

\section{InterTrans}

InterTrans is an interlinear text editor.  Interlinear text is a kind of
text in which each word is annotated with some combination of
phonological, morphological and
syntactic information (displayed under the word) and each sentence is
annotated with a free translation.  For an extended discussion of
interlinear text, and how to model it using annotation graphs,
see \cite{MaedaBird00}.

\begin{figure}[htbp]
  \begin{center}
    \leavevmode
    \epsfig{figure=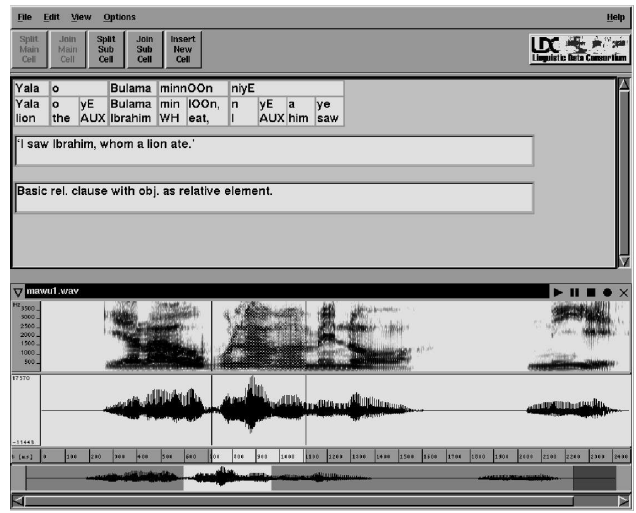, width=1.00\linewidth}
  \end{center}
  \caption{InterTrans, Interlinear Text Transcription}
  \label{fig:InterTrans}
\end{figure}

InterTrans permits interlinear transcription aligned to a primary
audio signal, for greater convenience, accuracy and accountability.
Whole words and sub-parts of words can be easily aligned with the
audio.  Clicking on any part of the annotation causes the corresponding
extent of audio signal to be highlighted.  As an extended recording is
played back, annotated sections are highlighted (both waveform and
interlinear text displays).  Figure~\ref{fig:InterTrans} contains a
screenshot of InterTrans.

The linguistic levels of information such as \emph{Free Translation (FT)},
\emph{Word} (\texttt{WD}) and \emph{Morphome} (\texttt{MP}) are called
\emph{types}.  The relationships among types are defined in a configuration
file.  For example, if the type \texttt{WD} is defined to \emph{dominate}
the type \texttt{MP}, operations described in this section, such as split
and join, on any cell of the type \texttt{WD} will cause corresponding cell
of the type \texttt{MP} to split or join as well.  It is also possible to
define two or more types as in an equivalent class.  In the implementation,
this is done by using multi-attribute labels of arcs.  For example, if the
types \emph{Morpheme} and (\texttt{MP}) and \emph{Morphemic Gloss}
(\texttt{MP-GLOSS}) are to be used, these can be defined as types of an
equivalent class.  Then, operations, such as \emph{Split} on any of the
cells of the type \texttt{MP} will cause the corresponding cell of the type
\texttt{MP-GLOSS} to split as well.

\paragraph{Insert New Cell}

A new cell can be inserted after the current cell by clicking on the
\emph{Insert New Cell} button.  A normal procedure for creating interlinear
text is to use this function first, and then split the sub cells as
necessary.

\paragraph{Delete Cell}

Deleting a cell can be done in one of the following ways after selecting 
the current cell; use the \emph{Control-d} key combination, or click
on the \emph{Delete Cell} button.

\paragraph{Split Cell}

Splitting a cell in InterTrans can be done in a similar manner to
MultiTrans.  First, the current cell should be selected.  Optionally, the
splitting point in the waveform can be chosen if the current cell is
associated with a region in the waveform.  Pressing the \emph{Return} key,
or clicking on the \emph{Split Cell} button will split the current
cell.\footnote{We plan to modify the program so that key bindings can be
  assigned by the user.}

\paragraph{Join Cell}

Joining two adjoining cells is simple.  First, the current cell should be
chosen.  Then clicking on the \emph{Join Sub Cell} button or pressing the
\emph{Control-j} key combination will join the current cell and its
preceding cell.

\paragraph{Alignment with Audio Waveform}

A partial or complete set of cells can be aligned with the audio signal
displayed in the waveform panel.  This can be done by selecting the
current cell in the interlinear transcription panel, and selecting the
corresponding region in the waveform panel.  The alignment can be changed
later.

\begin{figure*}[htbp]
  \begin{center}
    \leavevmode
    \epsfig{figure=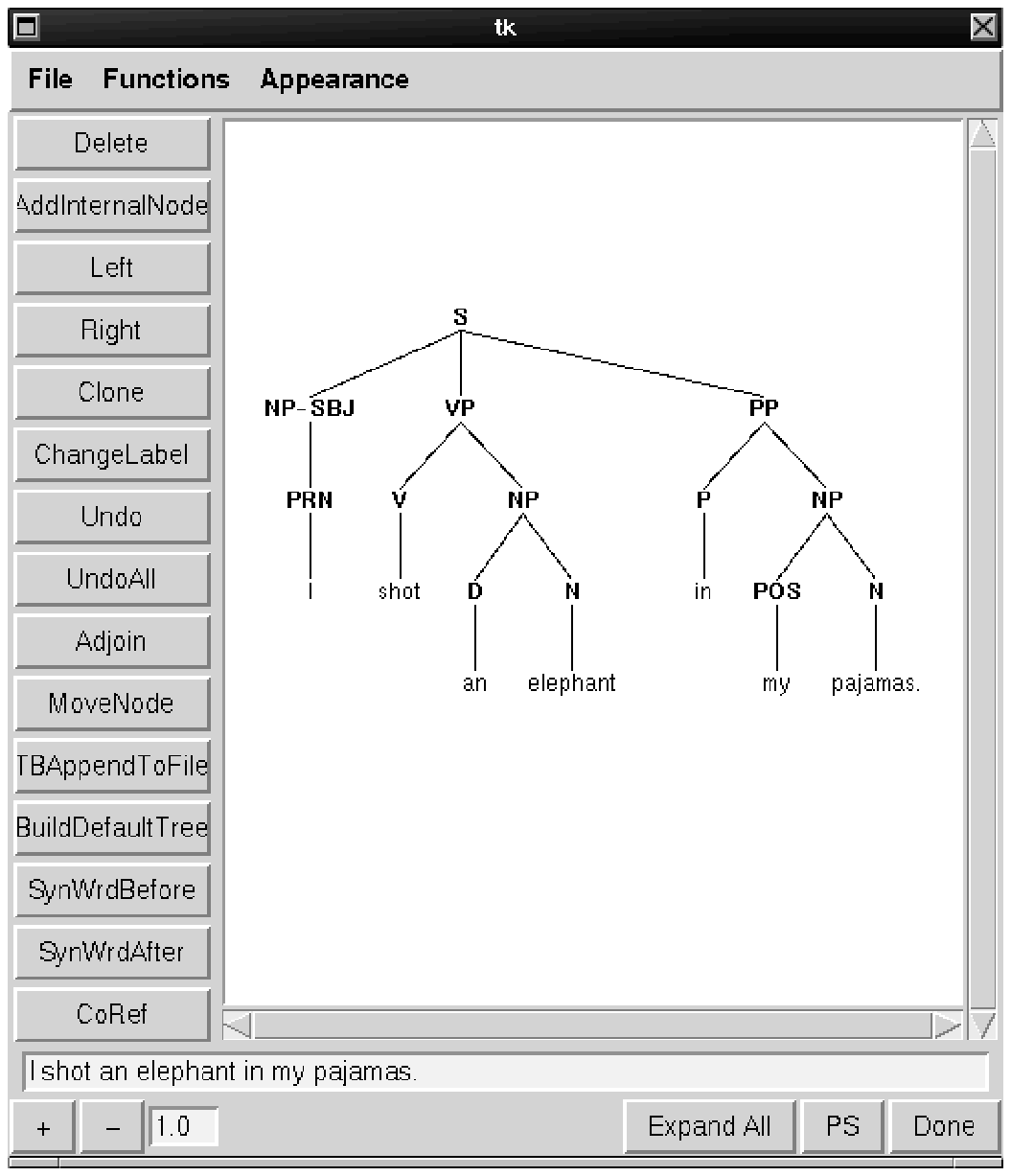, width=0.45\linewidth}
    \hspace{10mm}
    \epsfig{figure=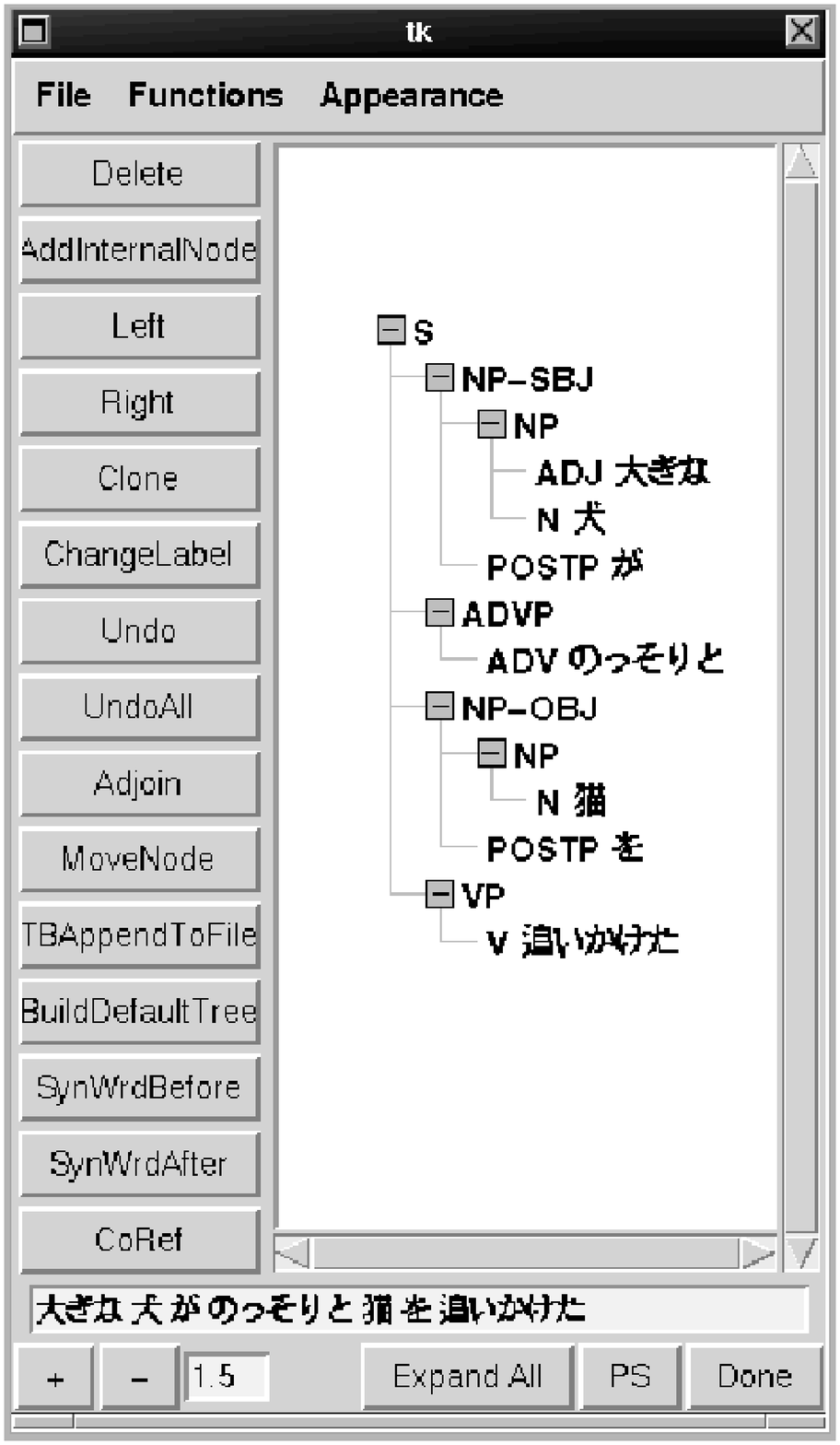, width=0.32\linewidth}
  \end{center}
  \caption{TreeTrans: Syntactic Transcription (Horizontal and Vertical Views)}
  \label{fig:TreeTrans}
\end{figure*}

\pagebreak

\section{TreeTrans}

TreeTrans is a tool for displaying and annotating tree structures.\footnote{
For a discussion of how to model trees using annotation graphs, see
\cite{BirdLiberman01,CottonBird02}.}
Basic annotation functions are provided, and functions can be
easily added or changed for specific annotation projects.  Trees
can be displayed in three forms: bottom-up, top-down, and vertical.
Two views of trees in TreeTrans are shown in Figure~\ref{fig:TreeTrans}.
Every node of the displayed tree has one of three
types: \texttt{syn}, a syntactic or internal node; \texttt{pos},
a part-of-speech node; or \texttt{wrd}, a node containing original text.

To change a displayed tree, users highlight the appropriate node(s)
then choose a function from the list of buttons. All the tree manipulation
functions preserve the surface order of the word string.

\paragraph{Input/Output}

TreeTrans can input either Penn Treebank-style bracketed trees or
annotation graphs.  Likewise, it can store annotated trees in either
of these formats.

\paragraph{InsertInternalNode}

If one node is highlighted, this function adds an internal node as the
parent of the highlighted node.  If two nodes are highlighted, this
function adds an internal node as the parent of the
the two highlighted nodes and any intervening material.
If the branches of the new internal node would cross any existing
branches of the tree, the insertion is not performed.

\paragraph{Delete}

This deletes the highlighted node.  If the highlighted node has type
\texttt{pos}, its corresponding \texttt{wrd} node is
also deleted.  If the deletion would cause an internal node to have no
leaves, the deletion is not performed.

\paragraph{MoveNode}

When two nodes are highlighted, this moves the first highlighted node and
its subtree to become a child of the second highlighted node.  When three
nodes are highlighted, the first two nodes and
any intervening material are moved to become children of the third
highlighted node.
The move is not performed if it would cause word order to be changed,
or if it would result in an internal node with no leaves.

\paragraph{Adjoin}
This is a syntactic adjoin, which creates a clone of the second highlighted
node, then moves the first highlighted node to become a child of the clone.

\paragraph{SynWrdBefore/After}

This function adds a pair of nodes, one of type \texttt{syn}, and one of type
\texttt{wrd}.  It is used for adding traces to the text. The new \texttt{syn}
node is a previous or following sibling of the highlighted node.

\paragraph{ChangeLabel}

This changes the label of the highlighted node.

\paragraph{CoRef}

This gives two highlighted nodes the same trace number.

\paragraph{BuildDefaultTree}

This builds a basic tree structure using the input sentence
as the terminals of the tree.
\arXivhack

\section{Obtaining the tools}

All of the annotation tools described in this paper are distributed under
an open source license, and available from
\smurl{http://www.ldc.upenn.edu/AG/}.

\bibliographystyle{lrec2000k}

\end{document}